\documentclass[11pt,a4paper]{article}
\usepackage[hyperref]{emnlp2018}
\usepackage{times}
\usepackage{latexsym}

\usepackage{graphicx}
\usepackage{amsfonts}
\usepackage{amssymb}
\usepackage{amsmath}

\usepackage{url}

\usepackage[disable]{todonotes} % add [disable] before {todonotes} to disable
\aclfinalcopy % Uncomment this line for the final submission
\title{\textit{Indicatements} that character language models learn English morpho-syntactic units and regularities}

\author{Yova Kementchedjhieva \\
  University of Copenhagen \\
  {\tt yova@di.ku.dk} \\\And
  Adam Lopez \\
  University of Edinburgh \\
  {\tt alopez@inf.ed.ac.uk} \\}

\date{}

\begin{document}
\maketitle
\begin{abstract}
%   We presents the results of an exploratory study into the morpho-syntactic awareness of character-level neural language models. We analyze an English character-to-character language model in isolation and as embedded into systems for morphological and syntactic analysis. Findings point to an ability of the model to identify units of higher order, such as morphemes and words, when an explicit cue is available. For derivational suffixes, in particular, we find that the model is able to learn the selectional restrictions of those with respect to their base, which suggests that character-level language models can capture subword-level linguistic regularities at the intersection of morphology and syntax.
% * <adam.d.lopez@gmail.com> 2018-07-10T10:10:21.441Z:
%
% ^.
Character language models have access to surface morphological patterns, but it is not clear whether or \textit{how} they learn abstract morphological regularities. We instrument a character language model with several probes, finding that it can develop a specific unit to identify word boundaries and, by extension, morpheme boundaries, which allows it to capture linguistic properties and regularities of these units. Our language model proves surprisingly good at identifying the selectional restrictions of English derivational morphemes, a task that requires both morphological and syntactic awareness. Thus we conclude that, when morphemes overlap extensively with the words of a language, a character language model can perform morphological abstraction.     
\end{abstract}

\section{Introduction}
%Character-level processing is now a common approach in speech recognition \citep{chan2016listen} and machine translation \citep{weiss2017sequence,Lee2016FullySegmentation,chung2016character}. 
%The increase in performance observed in these establishing works is speculatively attributed to several factors: more appropriate handling of words that were not seen during training; shared statistics between rare words and related more frequent words; and, importantly, linguistic awareness on the subword level.
%\yova{should maybe site specific papers?}  
%Research into the benefits of subword-level processing has found that the typological properties of languages do interact with the performance of subword models \citep{Vania2017From}. Yet, the exact aspects of this interaction are still to be surveyed in depth.\par 
Character-level language models \citep{sutskever2011generating} are appealing because they enable open-vocabulary generation of language, and \emph{conditional} character language models have now been convincingly used in speech recognition \citep{chan2016listen} and machine translation \citep{weiss2017sequence,Lee2016FullySegmentation,chung2016character}. They succeed due to parameter-sharing between frequent, rare, and even unobserved training words, prompting claims that they learn morphosyntactic properties of words. For example, \citet{chung2016character} claim that character language models yield ``better modelling [of] rare morphological variants'' while \citet{kim2016character} claim that ``Character-level models obviate the need for morphological tagging or manual feature engineering.'' But these claims of morphological awareness are backed more by intuition than direct empirical evidence. What do these models really learn about morphology? And, to the extent that they learn about morphology, \emph{how} do they learn it?

Our goal is to shed light on these questions, and to that end, we study the behavior of a character-level language model (hereafter LM) applied to English.
%, with the aim to identify its source of performance and to draw conclusions about theoretical limitations and venues for further improvements. 
We observe that, when generating text, the LM applies certain morphological processes of English productively, i.e. in novel contexts (\textsection 3). This rather surprising finding suggests that the model can identify the morphemes relevant to these processes. An analysis of the LM's hidden units presents a possible explanation: there appears to be one particular unit that fires at morpheme and word boundaries (\textsection 4). 
%Embedding the LM into a system for morphological segmentation
Further experiments reveal that the LM learns morpheme boundaries through extrapolation from word boundaries (\textsection 5).
%---this finding has strong implications for languages other than English, where the overlap between morpheme and word boundaries may not be so extensive.  
In addition to morphology, the LM appears to encode syntactic information about words, i.e. their part of speech (\textsection 6). With access to both morphology and syntax, the model should also be able to learn linguistic phenomena at the intersection of the two domains, which we indeed find to be the case: the LM captures the (syntactic) selectional restrictions of English derivational morphemes, albeit with some incorrect generalizations (\textsection 7).     
The conclusions of this work can thus be summarized in two main points---a character-level language model can:
\begin{enumerate}
\item learn to identify linguistic units of higher order, such as morphemes and words.
\item learn some underlying linguistic properties and regularities of said units.
\end{enumerate}
%The research questions we address are (a) does a character-level language model internalize linguistic structure beyond surface orthography, (b) does it encode any morphological information, (c) does it encode any syntactic information, and (d) can it learn linguistic regularities at the interface of morpho-syntax? Our results suggest that our character-level language model:
%\begin{itemize}
%\item developed a specific unit that recognizes words and, consequently, it can recognize morphemes as well, when those are word-forming or word-final at least some of the time.
%\item  can learn the linguistic regularities concerning identifiable morphemes, one example being the selectional restrictions of derivational suffixes in English, without any explicit morpho--syntactic information.
%\end{itemize}

%These conclusions were drawn based on (a) a preliminary analysis of the trained language model, on experiments with embedding the language model into (b) a system for morphological segmentation, and (c) a system for part-of-speech tagging, and finally, on (d) experiments probing the model for awareness with respect to specific morphological regularities. The following section provides background on each of these four parts of the study, 
%highlighting their properties as a diagnostic for linguistic awareness in the language model
%. The rest of the paper addresses the methods, results and conclusions from each part individually. 

\section{Language Modeling}

The LM explored in this work is a `wordless' character RNN with LSTM units \citep{karpathy2015unreasonable}.\footnote{\citet{karpathy2015unreasonable} is a blog post that discusses exactly this model. We are unaware of scholarly publications that use this model in isolation, though it is used in several conditional models \citep{chan2016listen,weiss2017sequence,Lee2016FullySegmentation,chung2016character}, and it is similar to the character RNN of \citet{sutskever2011generating}, which uses a multiplicative RNN rather than an LSTM unit.} It is `wordless' in the sense that input is not segmented into words, and spaces are treated just like any other character. This architecture allows for experiments on a subword, i.e. morphological level: we can feed a partial word and ask the model to complete it or record the probability the model assigns to an ending of our choice.\par

%\begin{figure}
%	\centering
%    \includegraphics[width=8cm]{the LM.pdf}
%    \caption{the LM Model Architecture.}
%    \label{fig:the LM}
%\end{figure}

\subsection{Formulation}
At each timestep $t$, character $c_{t}$ is projected into a high-dimensional space by a character embedding matrix $\textbf{E} \in \mathbb{R}^{|V|\times d}$: $\textbf{x}_{c_{t}} = \textbf{E}^{T} \textbf{v}_{c_t}$, where $|V|$ is the vocabulary of characters encountered in the training data, $d$ is the dimension of the character embeddings and $\textbf{v}_{c_t} \in \mathbb{R}^{|V|}$ is a one-hot vector with $c_t$th element set to 1 and all other elements set to zero.\par
%, where $i$ corresponds to the position of character  $c_{t}$ in the vocabulary, $V$. \par
The hidden state of the neural network is obtained as: $\textbf{h}_{t} = \operatorname{LSTM} (\textbf{x}_{c_{t}};\textbf{h}_{t-1})$. 
This hidden state is followed by a linear transformation and a softmax function over all elements of $V$, which results in a probability distribution. 
\begin{equation*}
p(c_{t+1} = c \mid \textbf{h}_t) = \operatorname{softmax}(\textbf{W}_o\textbf{h}_t + \textbf{b}_o)_{i} \quad \forall c \in V
\end{equation*}
where $i$ is the index of $c$ in $V$.\par 
%The embedding matrix, $\textbf{E}$, and all weights and bias terms, including the LSTM ones, are learned during training.

\subsection{Training}
The model was trained on a continuous stream of the first 7M character tokens from the English Wikipedia corpus. Data was not lowercased and it was randomly split into training (90\%) and development (10\%). Following a grid search over hyperparameters on a subset of the training data, we chose to use one layer with a hidden unit size of 256, a learning rate of 0.003, minibatch size of 50, and dropout rate of 0.2, applied to the input of the hidden layer.

\par
%For the purposes of meaningful the LM embedding, which is highly contextual, a past context is necessary. Yet, GS segmentations are available for words in isolation only. The Wikipedia dump, being one of the largest corpora of raw English text, was used to extract contexts for every word, taking the 15 tokens that preceded the word on up to 15 of its appearances in the dump. The occurrence of a word in each of its contexts was then treated as a separate training instance. Unfortunately, about 20\% of the words in the GS dataset never appear in the Wiki dump -- these words had to be processed in isolation, i.e. without a preceding context. This method resulted in a dataset of size 8989 on the word level, and 44937 on the character level. Training, development and test set were obtained with a random 80-10-10 split.\par 

\section{The English dialect of a character LM}
In an initial analysis of the learned model we studied text generated with the LM and found that it closely resembled English on the word level and, to some degree, on the level of syntax.
\subsection{Words}
%A language model can be used for text generation, performed by initializing the state of a trained model at random and sampling from it for T timesteps. The output can then be evaluated for 'naturalness'---overall, on the sentence level, or on the word level. 
When sampled, the LM generates real English words most of the time, and only about 1 in every 20 tokens is a nonce word.\footnote{As measured by checking whether the word appeared in the training data or in the \textit{pyenchant} UK or US English dictionaries.}
\begin{table}
\centering
 \begin{tabular}{||l||} 
 %\hline
 %Nonce Words\\ [0.5ex] 
 %\hline
 \hline
 sinding, fatities, complessed \\
 breaked, indicatement\\
 applie, therapie\\
 knwotator, mindt, ouctromor\\ [1ex] 
 \hline
 \end{tabular}
 \caption{Nonce words generated with the LM through sampling.}
 \label{table:noncewords}
\end{table}
Regular morphological patterns can be observed within some of these nonce words (Table~\ref{table:noncewords}). The words \textit{sinding, fatities} and \textit{complessed} all seem like well-formed inflected variants of English-looking words. The forms \textit{breaked} and \textit{indicatement} show productive morphological patterns of inflection and derivation being applied to bases of the correct syntactic class, namely verbs. It happens that \textit{break} is an irregular verb and \textit{indicate} forms a noun with suffix \textit{-ion} rather than \textit{-ment}, but these are lexical rules that block the more regular inflectional and derivational rules the LM has applied. 
In addition to composing morphologically complex words, the LM also attempts to decompose them, as can be seen with the forms \textit{therapie} and \textit{applie}. Here the inflectional suffix \textit{-s} has been dropped, but the orthographic change associated with it has not been successfully reversed. 
Not all nonce words generated by the LM can be explained in terms of morphological productivity, however: \textit{knwotator, mindt}, and \textit{ouctromor} don't resemble any real morphemes and don't follow English phonotactics. These forms may be highly improbable accidents of the sampling process. \par

\begin{table}
\centering
\resizebox{\linewidth}{!}{
 \begin{tabular}{||l||} 
 \hline
 The novel regarded the modern Laboratory has a\\
 weaken-little director and many of them in 2012\\
 to defeat in 1973 - or eviven of Artitagements.\\
[1ex] 
 \hline
 \end{tabular}
}
\caption{A sentence generated with the LM through sampling.}
\label{table:sentence}
\end{table}

\subsection{Sentences}
Consider the sentence in Table~\ref{table:sentence} generated with the LM through sampling. The sentence could not be considered fluent or grammatical: there are no clear dependencies between verbs, subjects, and objects; it contains the nonce word \textit{eviven} and the novel and unlikely compound \textit{weaken-little}. Yet, some short-distance syntactic regularities can be observed. Articles precede adjectives and nouns but not verbs, prepositions precede nouns, and particle \textit{to} precedes a verb. On an even larger scale, the clause \textit{the novel regarded the modern Laboratory has a weaken-little director} is grammatical in terms of the order between parts of speech\footnote{That is, assuming that \textit{novel} is a noun in this context and \textit{weaken-little} is an adjective, by analogy with its second base. }. The sentence appears unnatural due to its odd semantics, but consider the following alternative choice of words for the same syntactic structure: \textit{the man thought the modern laboratory has a weaken-little director}. This sentence sounds only marginally anomalous. \par 
The predominantly well-formed output of the LM suggests that it is appropriate to further study the linguistic regularities learned by it.\par 

\section{Meaningful hidden units in the LM}

\renewcommand{\arraystretch}{0.8}
\begin{table}
\centering
 \begin{tabular}{||c r||} 
 \hline
 Unit & $t_{-13}\qquad \ldots \qquad t$\\ [0.5ex] 
 \hline\hline
 \textit{Punctuation}&\texttt{r ( 1936-1939)}\\
  &\texttt{chool in 1921.}\\
  &\texttt{il 13 , 1813 .}\\
  &\texttt{ified in 1901.}\\
  &\texttt{\_( 1993-1998 )}\\
  \hline
 \textit{Word}&\texttt{'s predictions}\\
 &\texttt{ral relativism}\\
 &\texttt{\_contributions}\\
 &\texttt{\_were contract}\\
 &\texttt{at connections}\\
 \hline
\textit{Latinate suffix}&\texttt{ered in inform}\\
&\texttt{\_the concentra}\\
 &\texttt{ultural recrea}\\
 &\texttt{\_was accommoda}\\
 &\texttt{\_Reyes introdu}\\[1ex] 
 \hline
 \end{tabular}
 \caption{Top 5 Contexts for Three Units in the Network of the LM. The last character in each string marks the peak in activation.}
 \label{table:units}
\end{table}

%A language model can also be used as a text processing tool, when probed with existing text. In this setting we can either focus on the probability it assigns to a substring, or on the activations of the model's units in response to a substring. The latter method in particular is useful as a tool to interpret the inner workings of any neural model. 
The hidden units of the LM were analyzed by feeding the training data back into the system and tracking unit activations on each timestep, i.e. after every character. For each unit, the five inputs which triggered highest activation (highest absolute value) were recorded \citep{Kadar}.
About 40 units exhibited patterns of activation that could be identified as meaningful with the human eye. We selected three of the more interesting units to briefly discuss here. Table \ref{table:units} shows a list of the top five triggers for each of these units, together with up to 13 characters that preceded them, to put them in context. One unit, which we'll dub the \textit{punctuation} unit, seems to respond to closing punctuation marks. Another, dubbed the \textit{Latinate suffix} unit, appears to recognize contexts that are likely to precede suffix \textit{-ion} and its variants, \textit{-ation}, \textit{-ction} and \textit{-tion}.

\subsection{The \textit{word} unit}
\begin{figure}
    \includegraphics[width=\linewidth]{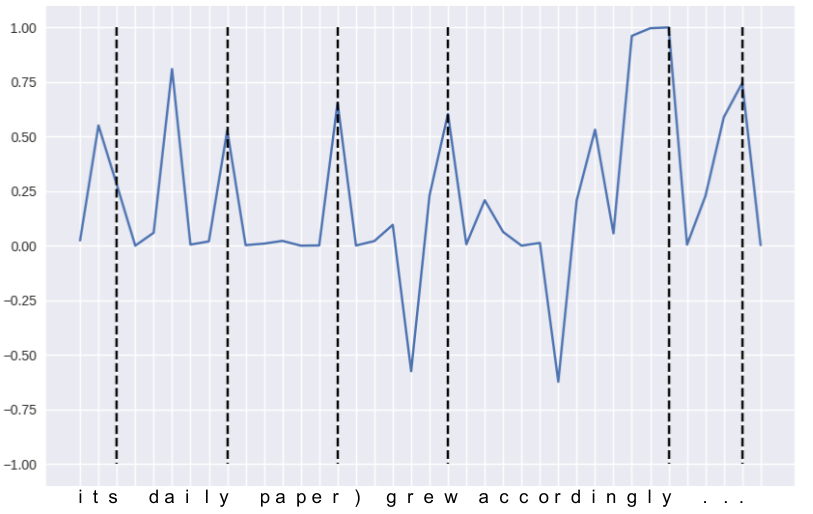}
    \caption{Activation of the \textit{word} unit. Query: \textit{its daily paper) grew accordingly}}
    \label{fig:unit13}
\end{figure}
The most interesting unit, the \textit{word} unit, appears to recognize complete words and sub-word units within them. Figure \ref{fig:unit13} shows the activation pattern of the \textit{word} unit over a partial sentence from the training data. The dotted lines, which mark the end of tokens, often coincide with the peaks in activation. The unit also recognizes the base \textit{it} within \textit{its} and \textit{according} within \textit{accordingly}. 
%Even the last dot in the ellipsis causes a high activation in the unit. 
The behavior of the unit could be explained either as a signal for the end of a familiar, repeated pattern, or as a predictor of a following space. The fact that we don't see a peak in activation at the right bracket symbol (which should be a cue for a following space) suggests that the former explanation is more plausible. Support for this idea comes from the low correlation coefficient between the activation of the unit and the probability the model assigns to a following space: only 0.08 across the entire training set. It appears that the LM knows that linguistic units need not occur on their own, i.e. that in a lot of cases a suffix is very likely to follow. \par

\section{Morphemes encoded by the LM}
Morphological segmentation aims to identify the boundaries in morphologically complex words, i.e. words consisting of multiple morphemes. A morpheme boundary could separate a base from an inflectional morpheme, e.g. \textit{like+s} and \textit{carrie+s}, a base from a derivational morpheme, e.g. \textit{consider+able}, or two bases, e.g. \textit{air+plane}. One approach to morphological segmentation is to see it as a sequence-labeling task, where words are processed one character at a time and every between-character position is considered a potential boundary. 
%The word $l_1i_2k_3e_4s$, for example, which has four between-character positions and a boundary in position 4, would be labeled $\langle$0001$\rangle$. 
RNNs are particularly suitable for sequence labeling and recent work on supervised morphological segmentation with RNNs shows promising results \citep{wang2016morphological}. \par
%The role of the RNN in this case, from an abstract point of view, is to encode the morphological structure of the input, such that the correct labels be assigned by the higher components of the system.

In this experiment we probe the LM using a model for morphological segmentation to test the extent to which the LM captures morphological regularities. \par
\subsection{Formulation}
At each timestep $t$, character $c_{t}$ is projected into high-dimensional space: \[\textbf{x}_{c_{t}} = \textbf{E}^{T} \textbf{v}_{i} \qquad  \textbf{E} \in \mathbb{R}^{|V_{char}|\times d_{char}}\] The hidden state of the encoder is obtained as before: $\textbf{h}^{enc}_{t} = LSTM^{enc} (\textbf{x}_{c_{t}};\textbf{h}^{enc}_{t-1})$. The hidden state of the decoder is then obtained as:
$\textbf{h}^{dec}_{t} = LSTM^{dec} (\textbf{h}^{enc}_{t};\textbf{h}^{dec}_{t-1})$
and followed by a linear transformation and a softmax function over all elements in $V_{lab}$, which results in a probability distribution over labels.
\begin{equation*}
 \begin{aligned}
 p(l_t = l \mid \textbf{c}, \textbf{l}_{word}) = \operatorname{softmax}(\textbf{W}^{dec}_o\textbf{h}^{dec}_t + \textbf{b}^{dec}_o)_i \\ \forall \quad l \in V_{lab} 
 \end{aligned}
\end{equation*}
where $\textbf{l}_{word}$ refers to all previous labels for the current word and $i$ refers to the index of $l$ in $V_{lab}$.\par 
The embedding matrix, $\textbf{E}$ and $LSTM^{enc}$ weights are taken from the LM. Decoder weights and bias terms are learned during training.\par

\subsection{Data}
The model (hereafter referred to as C2M, character-to-morpheme) was trained on a combined set of gold standard (GS) segmentations from MorphoChallenge 2010 and Hutmegs 1.0 (free data). The data consisted of 2275 word forms; 90\% were used for training and 10\% for testing. For the purposes of meaningful LM embeddings, which are highly contextual, a past context is necessary. Since GS segmentations are available for words in isolation only, we extract contexts for every word from the Wiki data, taking the 15 word tokens that preceded the word on up to 15 of its appearances in the dump. The occurrence of a word in each of its contexts was then treated as a separate training instance. \par 

\subsection{Performance}
The system achieved a rather low F1 score: 53.3. Compared to  \citet{ruokolainen2013supervised}, who obtain F1 score 86.5 with a bidirectional CRF model, C2M is clearly inferior. This is not particularly surprising given that the CRF makes predictions conditioned on past and future context, while C2M only has access to the past context. Recall also that the encoder of C2M shares the weights of the LM and is not fine-tuned for morphological segmentation. But taken as a probe of the LM's encoding, the F1 score of 53.3 suggests that this encoding still contains some information about morpheme boundaries. A breakdown of morphemic boundaries by type provides insights into the source of performance and limitations of C2M.\par 
\begin{table}
\centering
 \begin{tabular}{||l l l l||} 
 \hline
Model & Precision & Recall & F1\\[0.5ex] 
  \hline\hline
  C2M - WE &76.6&62.6&68.9\\
  C2M - $\neg$WE&23.1&34.2&27.6\\
  C2M - EOW&98.5&84.4&90.90\\
  C2M - $\neg$PREF&53.6&59.2&56.3\\[1ex] 
 \hline
 \end{tabular}
 \caption{C2M Performance. WE stand for word edge, EOW for end of word, and PREF for prefix.}
 \label{table:C2M}
\end{table}
\paragraph{Potential word endings as cues for morpheme boundaries}
The results labeled \textit{C2M - WE} and \textit{C2M - $\neg$WE} in Table \ref{table:C2M} refer to two types of morpheme boundaries: boundaries that could also be a  word ending (WE), e.g. \textit{drink+ing, agree+ment}, and boundaries that could not be a word ending ($\neg$ WE), e.g. \textit{dis+like, intens+ify}. It becomes apparent that a large portion of the correct segmentations produced by C2M can be attributed to an ability to recognize word endings. Earlier findings relating to the \textit{word} unit of the LM (section 4.1) align with this line of argument: the unit indeed detects words, and those morphemes that resemble words. 
The sample segmentations in A and C of Table \ref{table:segmentation} can be straightforwardly explained in terms of transfer knowledge on word endings: \textit{act, action} and \textit{ant} are all words the LM has encountered during training. Notice that the morpheme \textit{ant} has not been observed by C2M, i.e. it is not in the training data, but its status as a word is encoded by the LM. \par 

\begin{table*}
\centering
 \begin{tabular}{||l l l l l||} 
 \hline
&Input & True Segmentation & Predicted Segmentation&Correct\\[0.5ex] 
 \hline\hline
A.&actions&act+ion+s&act+ion+s&$\checkmark$\\
B.&acquisition&acquisit+ion&acquisit+ion&$\checkmark$\\
C.&antenna&antenna&ant+enna&\\
D.&included&in+clud+ed&in+clude+d&$\checkmark$\\
E.&intensely&in+tense+ly&intense+ly&\\
F.&misunderstanding&mis+under+stand+ing&misunder+stand+ing&\\
G.&woodwork&wood+work&wood+work&$\checkmark$\\[1ex] 
 \hline
 \end{tabular}
 \caption{Sample predictions of morphological segmentations.}
 \label{table:segmentation}
\end{table*}

\paragraph{Actual word endings}
An interesting result emerges when C2M's performance is tested on word-final characters, which by default should all be labeled as a morpheme boundary (\textit{C2M - EOW} in Table \ref{table:C2M}). Recall that the rest of the results exclude these predictions, since morphological segmentation concerns word-internal morpheme boundaries. C2M performs extremely well at identifying actual word endings. The margin between \textit{C2M - WE} results and \textit{C2M - EOW} results is substantial, even though both look at units of the same type, namely words. The higher accuracy in the EOW setting shows that the LM prefers to ends word where they actually end, rather than at earlier points that would have also allowed it. The LM thus appears to take into consideration context and what words would syntactically fit in it.
Consider example E in Table \ref{table:segmentation}. This instance of the word \textit{in+tense+ly} occurred in the context of \textit{the Himalayan regions of}. In this context C2M fails to predict a morpheme boundary, even though \textit{in} is a very frequent word on its own. The LM may be aware that preposition \textit{in} would not fit syntactically in the context, i.e. that the sequence \textit{regions of in} is ungrammatical. It thus waits to see a longer sequence that would better fit the context, such as \textit{intense} or \textit{intensely}. 
Example D shows that C2M is indeed capable of segmenting prefix \textit{in} in other contexts. The word \textit{included} is preceded by the context \textit{the English term propaganda}. This context allows a following preposition: \textit{the English term propaganda in}, so the LM predicted that the word may end after just \textit{in}, which allowed C2M to correctly predict a boundary after the prefix. \par 

%\paragraph{Bound Bases}
%Morpheme \textit{acquisit} (as in \textit{acquisition}, example B), was not observed by C2M during training and it is a bound morpheme that could not have been observed on its own by the LM. Yet, it was segmented out correctly on five out of fifteen occasions in the test data. We generated endings with the LM conditioned on \textit{acquisit} in its different contexts and found that the LM always predicted the expected ending \textit{ion}. It appears that C2M managed to generalize from free bases that take \textit{ion}, such as \textit{correct, react, predict} to the bound base \textit{acquisition}. Recall that the LM has a unit that specializes in recognizing bases that take suffix \textit{ion} (section 3.5.1). This explains the model's ability to share knowledge between different bases that all take that suffix.\par  
%\subsection{Conclusion}
%The main takeaway from the performance of C2M is that the LM is remarkably better at capturing word edges than morpheme boundaries, likely due to the explicit cues provided by the space character and punctuation. The fact that morpheme boundaries often resemble word edges in English allows the LM to indirectly capture some morphological boundaries such as those following a free base or a suffix. This finding explains the LM's ability to compose novel morphologically complex forms, such as \textit{sinding} and \textit{fatities}, and to decompose familiar ones. Yet, it also makes it theoretically impossible for the LM to segment out most prefixes, since they don't occur as independent words. \par 

\section{Parts of speech encoded by the LM}
Part-of-speech (POS) tagging is also seen as a sequence labeling task because words can take on different parts of speech in different contexts.
%Occasionally, new words would be encountered at test time, possibly in a context that is not sufficiently informative of their syntactic category. The context \textit{I saw a \_ rocket}, for example, could either take an adjective like \textit{big}, or a noun like \textit{space}. If the word \textit{misguided} happens to never appear in the training data used to learn a POS tagger, its tag could not unambiguously be determined from this context. 
Access to the subword level can be highly beneficial to POS tagging, since the shape of words often reveals their syntax: words ending in \textit{-ed}, for example, are much more often verbs or adjectives than nouns. Recent studies in the area of POS tagging demonstrate that processing input on a subword-level indeed boosts the performance of such systems \citep{dosSantos2014LearningTagging}.
Here we probe the LM with a POS-tagging model, hereafter C2T (character-to-tag). Its formulation is identical to that of C2M. 
\subsection{Data}
We used the English UD corpus (with UD POS tags) with an 80-10-10 train-dev-test split. Training data for character-level prediction was created by pairing each character of a word with the POS tag of that word, e.g. 'like$_{\textit{VERB}}$' was labeled as as $\langle$\textit{VERB} \textit{VERB} \textit{VERB} \textit{VERB}$\rangle$. UD doesn't specify a POS tag for the space character, so we used the generic X tag for it. Similarly to C2M, encodings for C2T were obtained for words in context.
\subsection{Performance}
C2T obtained an accuracy score of 78.85\% on the character level and 87.06\% on the word level, where word-level accuracy was measured by comparing the tag predicted for the last character of a word to the gold standard tag. The per-character score is naturally lower by a large margin, as predictions early on in the word are based on very little information about the identity of the word. Notice that the per-word score for C2T falls short of the state-of-the-art in POS tagging due a structural limitation: the tagger assigns tags based on just past and present information. The high accuracy of C2T in spite of this limitation suggests that the majority of the information concerning the POS tag of a word is contained within that word and its past context, and that the LM is particularly good at encoding this information.\par 
\begin{figure*}
    \includegraphics[width=16cm]{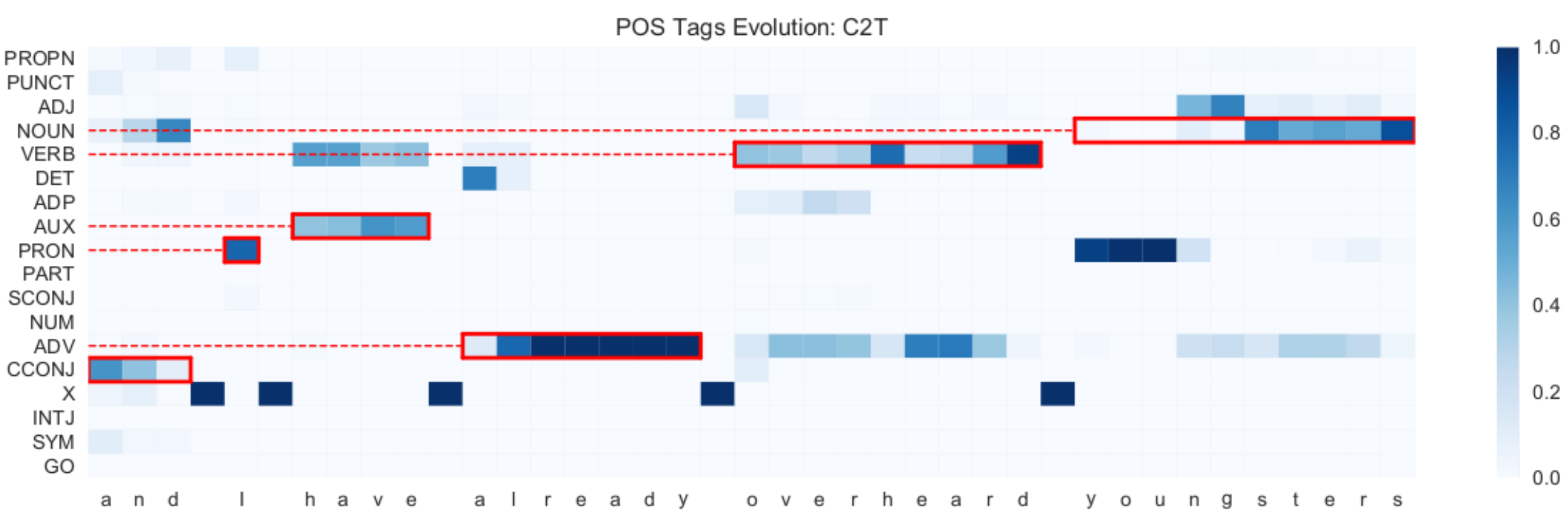}
    \label{fig:tag_evolution_perchar}
    \caption{C2T Evolution of POS Tag Predictions. Red rectangles point to the correct tags of words.}
\end{figure*}
\paragraph{Evolution of Tag Predictions over Time}
Figure \ref{fig:tag_evolution_perchar} illustrates the evolution of POS tag predictions over the string \textit{and I have already overheard youngsters} (extract from the UD data) as processed by C2T. 
Early into the word \textit{already}, for example, C2T identifies the word, recognizes it as an adverb and maintains this hypothesis throughout. With respect to the morphologically complex word \textit{youngsters} we see C2T making reasonable predictions, predicting PRON for \textit{you-}, ADJ for the next two characters and NOUN for \textit{youngster-} and \textit{youngsters}.

%\paragraph{Space characters}
%The fact that C2T assigned a PUNCT tag to some spaces means that on these occasions it recognized that the previous word had ended and that what it had seen since then (i.e. the space character) didn't constitute a word in itself, therefore it must be a punctuation mark. This observation provides further support for the idea that the LM sees words as units with an identifiable end.\par
%\subsection{Conclusion}
%The performance of the POS tagger based on the LM encodings demonstrated a high level of syntactic awareness in the model on a word-to-word level. This finding is in line with earlier observations from text generated by the LM.
%The model learns co-occurrences not just between words but between syntactic categories. This allows it to generate phrases that may be odd in meaning, like the one above, but mostly abide by the combinatorial properties of syntax within some temporal window.  \par 

\section{Selectional restrictions in the LM}
English derivational suffixes have \textbf{selectional restrictions} with respect to the syntactic category of the base they would attach to.  Suffixes \textit{-al, -ment} and \textit{-ance}, for example, only attach to verbs, e.g. \textit{betrayal, annoyance, containment}, while \textit{\mbox{-hood}}, \textit{\mbox{-ous}}, and \textit{-ic} only attach to nouns, as in \textit{nationhood, spacious and metallic}.\footnote{All examples are from \citet{77d0a4170a454dc0a724c220f7115f86}.} The former are thus known as \textbf{deverbal}, and the latter as \textbf{denominal}. Certain suffixes are members of more than one class, e.g. \textit{-ful} attaches to both verbs and nouns, as in \textit{forgetful} and \textit{peaceful}, respectively.  Since our LM appears to encode information about (some) morphological units and part of speech, it is natural to wonder whether it also encodes information about selectional restrictions of derivational suffixes. If it does, then the probability of a deverbal suffix should be significantly higher after a verbal base than after other bases, likewise with denominal suffixes and nominal bases. 
Our next experiment tests whether this is so.\par
\subsection{Method}
Our experiment measures and compares the probability of suffixes with different selectional restrictions across subsets of nominal, verbal,  and adjectival bases, as processed by the LM. We use carefully chosen nonce words as bases in order to abstract away from previously seen base-suffix combinations, which the model may have simply memorized.\par
\paragraph{Probability}
We compute the probability of a suffix given a base as the joint probability of its characters. For example, the probability of suffix \textit{-ion} attaching to base \textit{edit} is:
\[ p(\textit{ion} \mid \textit{edit}) = p(\textit{i} \mid \textit{edit}) \times p(\textit{o} \mid \textit{editi}) \times p(\textit{n} \mid \textit{editio})\]
Since the LM is a wordless language model, $p(\cdot|base)$ is approximated from $p(\cdot|\textbf{c})$ where $\textbf{c}$ is the entire past. The probability of a suffix in the context of a particular syntactic category was computed as the average probability over all bases belonging to that category.\par
\paragraph{Nonce Bases}
Nonce bases were obtained by sampling complete words from the LM---that is, sequences delimited by a space or punctuation on both sides. We discarded all words that appeared in an English dictionary, and imposed several restrictions on the remaining candidates: their probability had to be at most one standard deviation below the mean for real words (to ensure they weren't highly unlikely accidents of the sampling procedure), and the probability of a following space character had to be at most one standard deviation below the mean for real words (to avoid prematurely finished words, such as \textit{measu} and \textit{experimen}). In addition, nonce words had to be composed entirely of lowercase characters and couldn't end in a suffix (as certain suffixes included in the experiment only attach to base stems). The candidates that met these conditions were labeled for POS using C2T. The final nonce bases used were the ones whose POS tag confidence was at most one standard deviation below the mean confidence with which tags of real words were assigned. Some examples of nonce words from the final selection are shown in Table \ref{table:words}.  An embedding was recorded for every nonce word that met these conditions by taking the hidden state of the language model at the end of the word in context. \par
\begin{table}
\centering
\resizebox{\linewidth}{!}{
 \begin{tabular}{||l l||} 
\hline
Noun&crystale, algoritum, cosmony, landlough\\
Verb&underspire, restruct, actrace\\
Adjective&nucleent, transplet, orthouble\\ [1ex] 
 \hline
 \end{tabular}
}
\caption{Sample Nonce Bases}
\label{table:words}
\end{table}

\paragraph{Suffixes}
\begin{table}
\centering
\resizebox{\linewidth}{!}{
 \begin{tabular}{||l l||} 
 \hline
Noun&-ous, -an, -ic, -ate, -ary, -hood, -less, -ish\\
Verb&-ance, -ment, -ant, -ory, -ive, -ion, -able, -ably\\
 Adjective&-ness, -ity, -en\\ [1ex] 
 \hline
 \end{tabular}
}
\caption{Syntactically unambiguous derivational suffixes}
\label{table:suffixes}
\end{table}
The suffixes included in this experiment (listed in Table \ref{table:suffixes}) were taken from \citet{77d0a4170a454dc0a724c220f7115f86}, one of the most extensive studies of the selectional restrictions of English derivational suffixes. Fabb discussed 43 suffixes, many of which attach to a base of two out of three available syntactic categories, e.g. \textit{-ize} attaches to both nouns, as in \textit{symbolize}, and adjectives, as in \textit{specialize}. The analysis of such syntactically ambiguous suffixes is complex since the frequency with which they attach to each base type should be taken into consideration, but such statistics are not readily available and require morphological parsing. For the purposes of the present study ambiguous suffixes were thus excluded and only the remaining nineteen suffixes were used. \par 
\subsection{Results}
\begin{figure}
    \includegraphics[width=\linewidth]{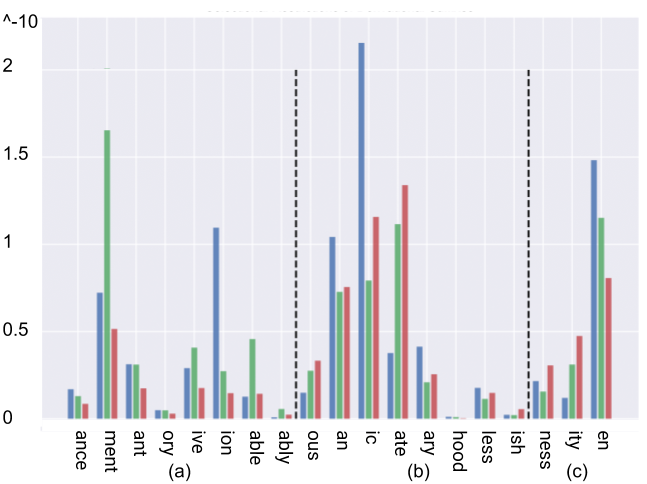}
    \caption{Suffix Probability Following Nominal (blue), Verbal (green) and Adjectival (red) Bases. Suffixes are grouped according to their selectional restrictions: (a) deverbal, (b) denominal and (c) deadjectival. 
	%Naive suffix frequency counts are reported at the  	bottom of the plot. 
	Eleven out of nineteen suffixes obtained highest probability following the syntactic category that matched their selectional restrictions.}
    \label{fig:suffix_exp}
\end{figure}
Figure \ref{fig:suffix_exp} shows the results from the experiment. Eleven out of nineteen suffixes exhibit the expected behavior: suffixes \textit{-ment, -ive, -able, -ably, -an, -ic, -ary, -hood, -less, ness} and \textit{-ity} are more probable in the context of their corresponding syntactic base than in other contexts. Suffix \textit{-ment}, for instance, is more than twice as probable in the context of a verbal base than in the context of a nominal or an adjectival base. The fact that almost 70\% of suffixes `select' their correct bases, points to a linguistic awareness within the LM with respect to the selectional restrictions of suffixes. 
%Had there been no such awareness, the ratio of success should have been close to 33\%. 
%\adam{I remove this last bit b/c I think a precise baseline would depend on frequency of category.}
\par 
Despite the overall success of the LM in this respect, some suffixes show a definitive preference for the wrong base. A further analysis of some of these cases shows that they don't necessarily counter the evidence for syntactic awareness within the LM. \par
\subsection{Suffix \textit{-ion}}
Deverbal suffix \textit{-ion} should be most probable following verbs, but prefers nominal bases. Notice that \textit{-ion} is a noun-forming suffix: \textit{communicate}$_{VERB}$ $\rightarrow$ \textit{communication}$_{NOUN}$, \textit{regress}$_{VERB}$ $\rightarrow$ \textit{regression}$_{NOUN}$. It appears that from the perspective of the LM, the syntactic category of such morphologically complex forms extends to their bases, e.g. the LM perceives the \textit{populat} substring in \textit{population} as nominal. This observation can be explained precisely with reference to the high frequency of suffix \textit{-ion}: the suffix itself occurred in 18,945 words in the dataset, while the frequency of its various bases in isolation, e.g. of \textit{populate, regress}, etc., was estimated to be only 9,755\footnote{To estimate this, we removed the suffix of each word and then searched for the remainder in isolation, followed by \textit{e} and followed by \textit{s/ es}---e.g. for \textit{population} we counted the occurrences of \textit{populat, populate, populates} and \textit{populated}.}. This shows that bases that can take suffix \textit{-ion} were seen more often with it than without it. As a consequence, the LM is biased to expect a noun when seeing one of these character sequences and may thus perceive the base itself as a nominal one. \par 

\begin{figure*}
	\centering
    \includegraphics[width=16cm]{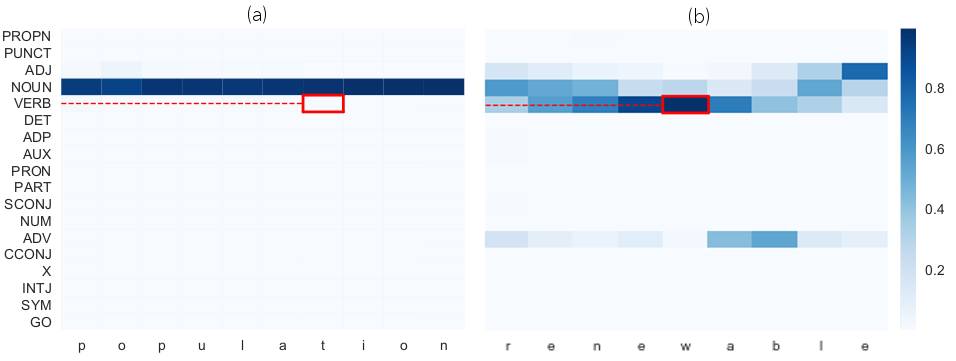}
    \caption{C2T Evolution: (a) \textit{population} and (b) \textit{renewable}. The red square points to the syntactic category of the base.}
    \label{fig:renewable}
\end{figure*}

The tag prediction evolution over \textit{population} and \textit{renewable} in Figure \ref{fig:renewable}, show that this is indeed the case, by comparing a base suffixed with \textit{-ion} to a base suffixed with \textit{-able} (whose selectional restrictions, we know, were learned correctly). For both words C2T starts off predicting NOUN. For \textit{renew} it switches to VERB, which is the correct tag for this base, and only upon seeing the suffix, progresses to the conclusive ADJ tag for \textit{renewable}. For \textit{population}, in contrast the prediction remains constant at NOUN, which is indeed the category of the word, as determined by the suffix.\par
%\begin{figure}
%	\centering
%    \includegraphics[width=\linewidth]{POS_tag_evolution_population.pdf}
%    \caption{C2T Evolution: \textit{population}. The red square points to the syntactic category of the base.}
%    \label{fig:manipulate}
%    \includegraphics[width=\linewidth]{POS_tag_evolution_renewable.png}
%    \caption{C2T Evolution: \textit{renewable}. The red square points to the syntactic category of the base.}
 %   \label{fig:renewable}
%\end{figure}

%\section{Conclusion}
%The experiment demonstrated that the LM is capable of capturing the selectional restrictions of derivational suffixes with respect to the syntactic category of bases, when there are no strong interfering factors. This outcome is in line with the findings of earlier chapters: a language model that is capable of encoding information concerning the syntactic and morphological structure of its input (here, suffixed words), is also capable of learning morpho-syntactic dependencies in the input.\par

\section{Conclusion}
This work presented an exploratory analysis of a `wordless' character language model, aiming to identify the morpho-syntactic regularities captured by the model.\par 
%\subsection{Summary of Results}
%A specific unit was identified that demonstrated the model could recognize words as separate units. More evidence in this respect was found in an experiment embedding the language model into a system for morphological segmentation. The experiment also showed that the model did not inherently identify morpheme boundaries, but it could transfer knowledge on word edges to the task of morphological segmentation, whenever conditions allowed it. Consequently, it encoded boundaries following a base or a suffix better than those following a prefix. \par 
%An experiment embedding the LM into a part-of-speech tagging model provided further evidence for awareness on the word level and showed that the model was capable of encoding syntactic information. \par
%Finally, a linguistic experiments tested the ability of the language model to learn a regularity concerning the selectional restrictions of derivational suffixes. The model performed mostly well on the experiment, with some exceptions being easily accounted for.\par
The first conclusion of this work is that morpheme boundaries are mainly learned by the LM through analogy with syntactic boundaries. 
Findings relating to the extremely frequent suffix \textit{-ion} illustrate that the LM was able to learn to identify purely morphological boundaries through generalization. But a prerequisite for this generalization is that a morpho-syntactic boundary was also seen in the relevant position during training. \par 

The second conclusion is that having recognized certain boundaries and by extension, the units that lie between them, the model could also learn the regularities that concern these units, e.g. the selectional restrictions of most derivational suffixes included in the study could be captured accurately.\par 

\section{Implications for future research}

The above conclusions have strong implications with respect to the use of character-level LMs for languages other than English. 

English is the perfect candidate for character-level language modeling, due to its fairly poor inflectional morphology. The nature of English is such that the boundary between a base and a suffix is often also a potential word boundary, which makes suffixes easily segmentable. This is not the case for many languages with richer and more complex morphology. Without access to the units of verbal morphology, it is less clear how the model would learn these types of regularities. This shortcoming should hold not just for the LM but for any character-level language model that processes input as a stream of characters without segmentation on the subword level. \par

This implication is in line with the results of \citet{Vania2017From} showing that for many languages, language modeling accuracy improves when the model is provided with explicit morphological annotations during training, with English showing relatively small improvements. Our analysis might explain why this is so; we expect analyses of other languages to yield further insight. \par

Finally, we should point out that it may not be the case that a single, highly-specified \textit{word} unit should exist in every character-level LM. \citet{qian2016analyzing} find that different levels of linguistic knowledge are encoded with different model architectures, and \citet{kadar2018revisiting} find that even a different initialization of an otherwise identical model can results in very different hierarchical processing of the input. We consider ourselves lucky for coming across this particular setup that produced a model with very interpretable behavior, but we also acknowledge the importance of evaluating the reliability of the \textit{word} unit finding in future work.

\section*{Acknowledgements}
We thank Sorcha Gilroy, Joana Ribeiro, Clara Vania, and the anonymous reviewers for comments on previous drafts of this paper.

\bibliography{irp,acl2018}
\bibliographystyle{acl_natbib_nourl}

\end{document}